\def\year{2022}\relax
\documentclass[letterpaper]{article} 
\usepackage{aaai22}  
\usepackage{times}  
\usepackage{helvet}  
\usepackage{courier}  
\usepackage[hyphens]{url}  
\usepackage{graphicx} 
\usepackage{natbib}  
\usepackage{caption} 
\DeclareCaptionStyle{ruled}{labelfont=normalfont,labelsep=colon,strut=off} 
\usepackage{algorithm}
\usepackage{algorithmic}

\usepackage{booktabs}
\usepackage{multirow}
\usepackage{amsmath}
\usepackage{subfigure}
\usepackage{newfloat}
\usepackage{listings}
\floatstyle{ruled}
\newfloat{listing}{tb}{lst}{}
\floatname{listing}{Listing}

\graphicspath{{picture/}}

\usepackage{xspace}

\makeatletter
\DeclareRobustCommand\onedot{\futurelet\@let@token\@onedot}
\def\@onedot{\ifx\@let@token.\else.\null\fi\xspace}

\def\eg{\emph{e.g}\onedot} 
\def\ie{\emph{i.e}\onedot}

\makeatother

\title{Topology-aware Convolutional Neural Network for Efficient Skeleton-based Action Recognition}
\author{
    Kailin Xu\textsuperscript{\rm 1}\footnotemark[1]\footnotemark[2],
    Fanfan Ye\textsuperscript{\rm 2}\footnotemark[2],
    Qiaoyong Zhong\textsuperscript{\rm 2},
    Di Xie\textsuperscript{\rm 2}\footnotemark[3]
}
\affiliations{
    \textsuperscript{\rm 1}Zhejiang University\quad
    \textsuperscript{\rm 2}Hikvision Research Institute\\
    kailinxu@zju.edu.cn, \{yefanfan,zhongqiaoyong,xiedi\}@hikvision.com
}

\begin{document}

\maketitle

\renewcommand{\thefootnote}{\fnsymbol{footnote}}
\footnotetext[1]{Work done as an intern at Hikvision Research Institute.}
\footnotetext[2]{Both authors contributed equally to this work.}
\footnotetext[3]{Di Xie is the corresponding author.}

\begin{abstract}
  In the context of skeleton-based action recognition, graph convolutional networks (GCNs) have been rapidly developed, whereas convolutional neural networks (CNNs) have received less attention. One reason is that CNNs are considered poor in modeling the irregular skeleton topology. To alleviate this limitation, we propose a pure CNN architecture named Topology-aware CNN (Ta-CNN) in this paper. In particular, we develop a novel cross-channel feature augmentation module, which is a combo of map-attend-group-map operations. By applying the module to the coordinate level and the joint level subsequently, the topology feature is effectively enhanced. Notably, we theoretically prove that graph convolution is a special case of normal convolution when the joint dimension is treated as channels. This confirms that the topology modeling power of GCNs can also be implemented by using a CNN. Moreover, we creatively design a SkeletonMix strategy which mixes two persons in a unique manner and further boosts the performance. Extensive experiments are conducted on four widely used datasets, \ie N-UCLA, SBU, NTU RGB+D and NTU RGB+D 120 to verify the effectiveness of Ta-CNN. We surpass existing CNN-based methods significantly. Compared with leading GCN-based methods, we achieve comparable performance with much less complexity in terms of the required GFLOPs and parameters. 
\end{abstract}

\section{Introduction}

\begin{figure}[t]
	\centering
	\includegraphics[width=\linewidth]{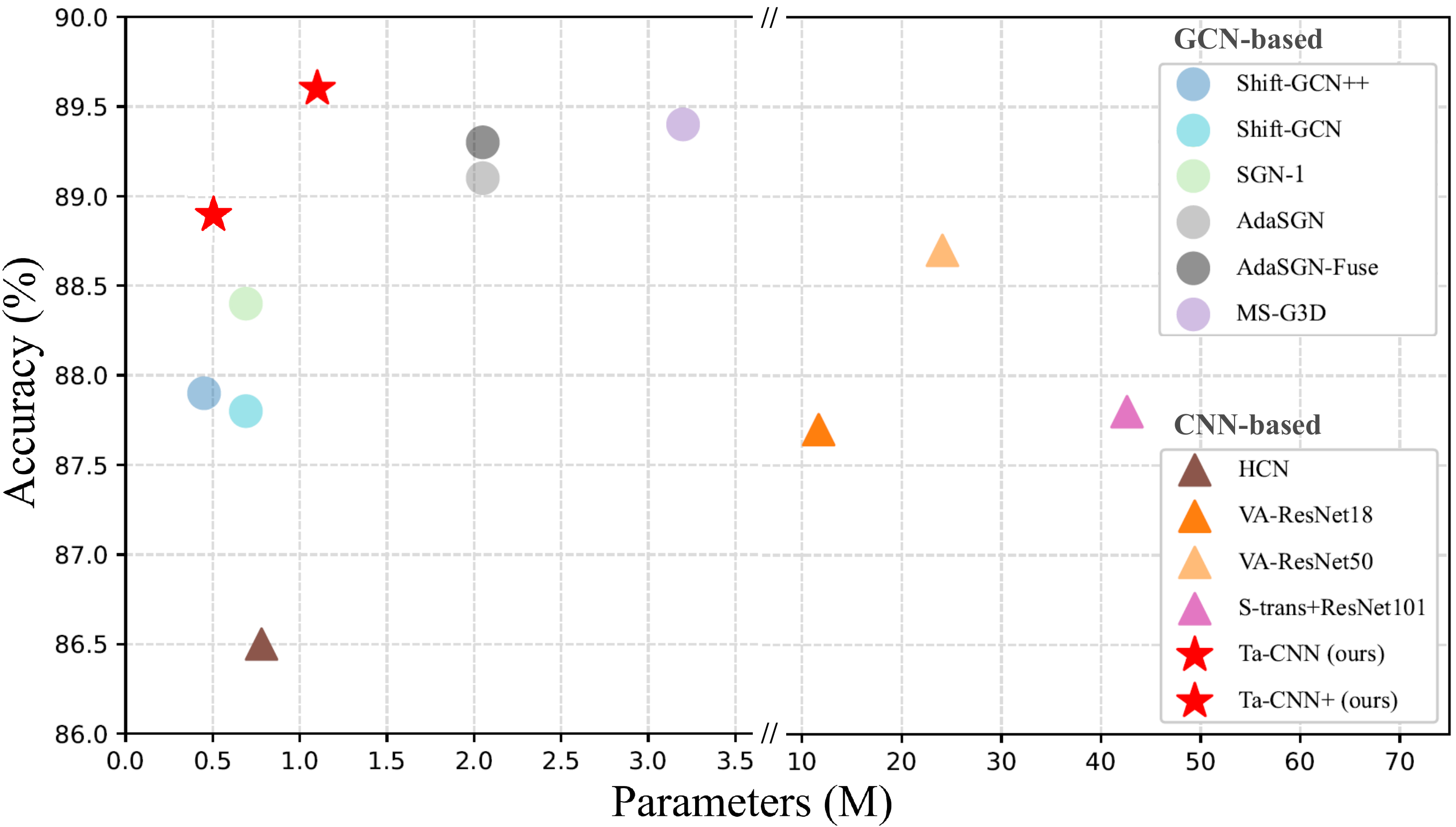} 
	\caption{Comparison of various methods on NTU RGB+D with the cross-subject benchmark in terms of accuracy and number of parameters. The proposed Ta-CNN achieves state-of-the-art performance with a tiny model size.}
	\label{compare}
\end{figure}

\begin{figure*}[t]
	\centering
	\includegraphics[width=0.92\linewidth]{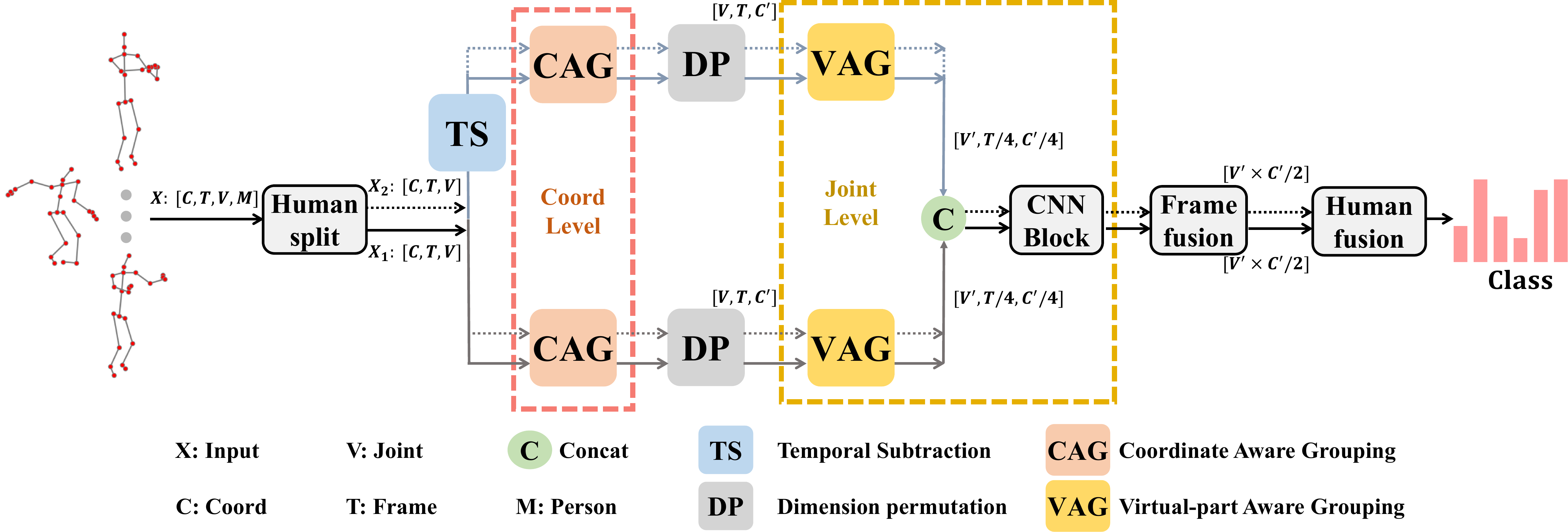} 
	\caption{Pipeline of the proposed Ta-CNN framework. Two streams, \ie the skeleton sequence and the motion information are fed into the network and fused in the middle. CAG and VAG are the key components to enhance the topology feature.}
	\label{pipline}
\end{figure*}

Skeleton-based action recognition has received widespread attention from the community, and many significant advances have been made in recent years. Unlike RGB-based image and video data, skeleton data are much more abstract and compact, reducing the parameters and computation resources required in the model design~\cite{zhang_semantics-guided_2020}. Looking back at the development of this field, methods that focus on hand-crafted features~\cite{vemulapalli_human_2014,zhang_view_2019} have been surpassed by deep learning-based methods. With the emergence of graph convolutional networks (GCNs)~\cite{yan_spatial_2018}, the recognition performance is being improved rapidly. GCN is considered to be able to handle the irregular topological information in skeleton data properly, which has been verified in many previous works~\cite{yan_spatial_2018,ye_dynamic_2020,chen_channel-wise_2021}. One obvious limitation is that most of these pure GCN-based methods rely on heavy models~\cite{liu_disentangling_2020,shi_adasgn_2021}. To attack the weakness, some works combine CNN and GCN into a compact model and get decent performance while balancing the model size~\cite{zhang_semantics-guided_2020}. However, few pure CNN-based methods~\cite{zhang_view_2019} were proposed to address the task and achieve competitive performance with GCN-based methods. It is widely recognized that CNN can not exploit the skeleton topology sufficiently. Inspired by this factor, a question came to our mind. Can we enhance the topology feature of a CNN model by borrowing the topology modeling insights from GCN-based methods?

Previous CNN-based methods mostly treat each dimension of the coordinate feature equally in convolutional layers. Also, none of them focuses on explicit modeling of the interaction among joints due to lack of a valid tool as the adjacency matrix for GCN. We argue that enhancing the coordinate and joint features is the key to effectively model the skeleton topology using CNN. To this end, we propose a novel Topology-aware CNN (Ta-CNN), which can fully mine the topological information of skeleton data and achieve remarkable performance. Based on CNN, it is easy to design a lightweight architecture with a low budget for GFLOPs and parameters (see Figure~\ref{compare}). This makes our model more friendly than GCNs for deployment.

The pipeline of the proposed Ta-CNN framework is illustrated in Figure~\ref{pipline}. Following HCN~\cite{li_co-occurrence_2018}, we first learn the per-joint coordinate feature, followed by per-person joint feature, and features of multiple persons are fused in a late stage. Specifically, we propose a novel cross-channel feature augmentation module, which is a combo of map-attend-group-map operations. It is adapted to enhance both coordinate feature and joint feature, resulting in the coordinate aware grouping (CAG) module and the virtual-part aware grouping (VAG) module. CAG is composed of three building blocks named feature mapping, channel attention and dual coordinate-wise convolution. This design is intuitive. In a multi-dimensional coordinate space, the importances of different coordinates are different for recognition of the underlying action. For example, it is easy to recognize the action of \emph{walking} from the side view, but difficult from the front view. To model the interaction among joints, we transpose the joint dimension into channels as is done in HCN. The transposition trick is the key to make pure CNN models benefit from the topology of skeleton data. With a theoretical analysis of convolution with the joint dimension as channels, we conclude that graph convolution can be regarded as a special case of convolution. To fully explore the topology within different body parts, we append the virtual-part aware grouping module after the CAG module. CAG and VAG effectively enhance the topology feature and thus boost the performance without introducing heavy extra computational cost.

In addition, inspired by the mixup~\cite{zhang_mixup_2018} strategy for augmentation of image data, we creatively invent a SkeletonMix strategy for augmentation of skeleton data. Due to the difference in modality, the vanilla mixup does not apply to skeleton data. In the proposed SkeletonMix, we mix two samples by randomly combining the upper body and the lower body of two different persons. SkeletonMix significantly enriches the diversity of samples during training and brings robustness and accuracy gain.

Our contributions can be summarized as follows:

\begin{itemize}
\item We propose a novel pure CNN-based framework for skeleton-based action recognition. Equipped with the CAG and VAG modules, the topology feature is effectively enhanced without requiring a heavy network.
\item We are the first to theoretically analyze and reveal the close relation between graph convolution and normal convolution, which may guide the design of CNN-based models.
\item A novel SkeletonMix strategy is invented for augmentation of skeleton data, leading to robust learning and better performance.
\item The proposed method surpasses all CNN-based methods and achieves comparable performance with GCN-based methods with significantly reduced parameters and GFLOPs.
\end{itemize}

\section{Related Work}
\paragraph{Skeleton-based Action Recognition.}
For this task, deep learning-based methods are the current state-of-the-arts and achieve significantly better performance than those using hand-crafted features~\cite{vemulapalli_human_2014,zhang_view_2019}. Here, we divide existing deep learning models into roughly three categories, namely RNNs, CNNs and GCNs.

RNN~\cite{Zaremba2014RecurrentNN} and its variants~\cite{cho_learning_2014} are a natural choice to model the temporal dynamics in skeleton data. In the early years, RNN has been incorporated for action recognition~\cite{shahroudy_ntu_2016,li_adaptive_2017}. Later many CNN-based methods emerged, because CNN is found to be able to encode spatiotemporal feature at the same time. To make CNN work on skeleton data, some works chose to convert skeleton information into images~\cite{du_skeleton_2015,liu_enhanced_2017,zhang_view_2019}, and others directly used the 3D skeleton data as input~\cite{kim_interpretable_2017,li_skeleton-based_2017,li_co-occurrence_2018}. \citeauthor{zhang_view_2019} combined the view adaptive module with CNN, achieving the best result among CNN-based methods.
	
Owing to the advantages in dealing with irregular graphical structures, GCNs have taken an absolute leading position in the context of skeleton-based action recognition. The first successful work is ST-GCN~\cite{yan_spatial_2018}, which learn the spatial and temporal patterns in skeleton data with graph convolution. Based on ST-GCN, there are many follow-up works which mainly focus on improving the local and global perception of GCN~\cite{liu_disentangling_2020,zhang_semantics-guided_2020,cheng_skeleton-based_2020,ye_dynamic_2020,chen_multi-scale_2021} and reducing the model complexity~\cite{zhang_semantics-guided_2020,cheng_skeleton-based_2020,ye_dynamic_2020}. The state-of-the-art performance is obtained by CTR-GCN~\cite{chen_channel-wise_2021}. This work proposed to relax the constraints of other graph convolutions and used the specific correlation to better model the channel topology. Although recent studies have mentioned the weakness of CNN in extracting skeleton topology information~\cite{yan_spatial_2018,ye_dynamic_2020,chen_channel-wise_2021, ye2019joints, ye2019skeleton}, in this paper, we prove that the potential of CNN has been underestimated. With a carefully designed feature augmentation module and training strategy, we are able to achieve competitive performance using a pure CNN model.

\paragraph{Mixup}
\cite{zhang_mixup_2018} is a data augmentation strategy which enriches the samples by mixing two images and their labels. It successfully improves the performance of state-of-the-art image recognition models on many datasets such as ImageNet and CIFAR. Then many variants of mixup are proposed by performing other types of mixing and interpolation~\cite{verma_manifold_2019}. \citeauthor{summers_improved_2019} studied a broader scope of mixed-example data augmentation. To address the generation of unnatural samples in mixup, CutMix~\cite{yun_cutmix_2019} replaces the image region with a patch cropped from another image. ResizeMix~\cite{qin_resizemix_2020} further improves CutMix by replacing the cropping operation with resizing of the whole source image. The above strategies can not be adopted for skeleton data directly, as they will cause severe unreasonable deformation to the skeleton topology. Instead, we invent a novel SkeletonMix strategy, which is specialized for augmentation of skeleton data and distinct from the way to mix images.

\section{Method}
In this section, we first briefly review the task of CNN-based skeleton-based action recognition. Then we elaborate the details of the proposed Ta-CNN framework, including the coordinate aware grouping module and the virtual-part aware grouping module. After that, the SkeletonMix strategy for skeleton data augmentation will be presented.

\subsection{Preliminaries}

Skeleton-based action recognition is essentially a classification problem. The input is a sequence of skeleton data, which can be denoted as $\mathbf{X} \in R^{C \times T \times V}$. $C$, $T$ and $V$ represent the number of coordinates, time sequence length and the number of joints respectively. In this form, a skeleton sequence can be interpreted as a multi-spectral image. That is, the coordinate dimension $C$ is analogous to the channels of an image, and the temporal and joint dimensions $T\times V$ can be regarded as height and width of the image. Following HCN~\cite{li_co-occurrence_2018}, the skeleton sequence is treated as a 3D tensor and fed into the network directly.

\subsection{Enhancing Coordinate Feature via CAG}

\begin{figure}[t]
	\centering
	\includegraphics[width=0.48\textwidth]{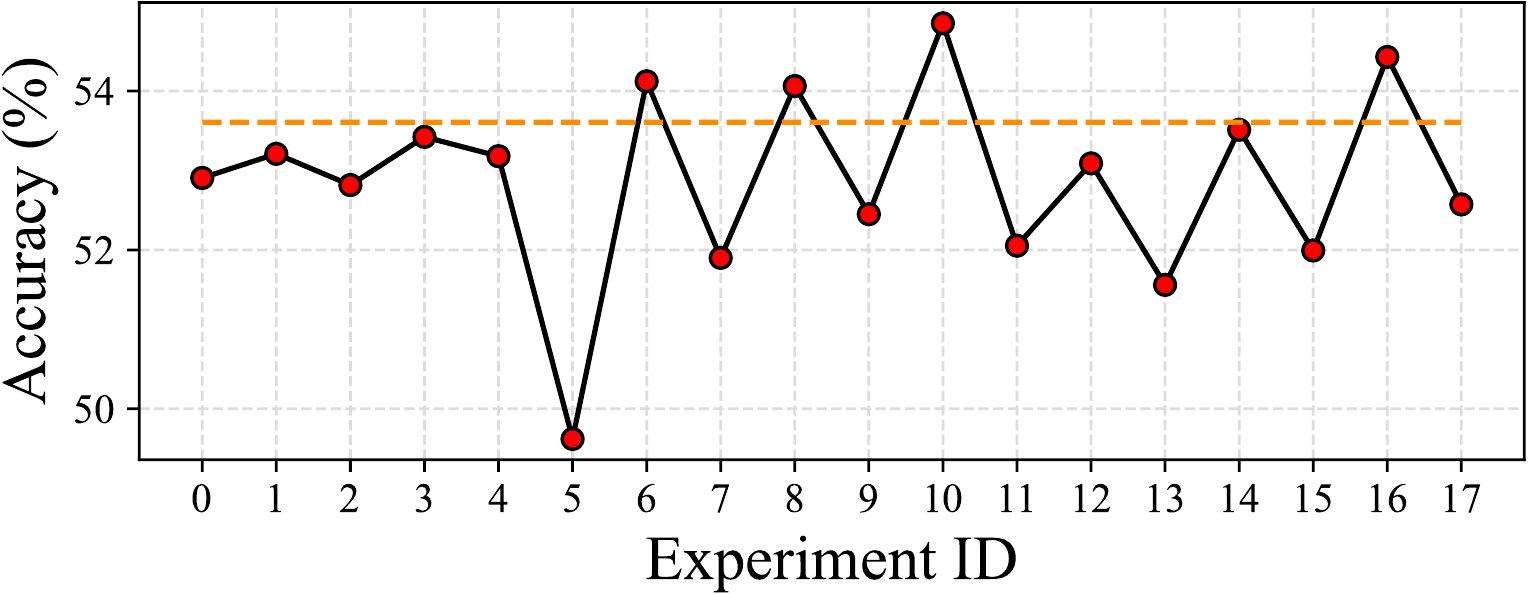} 
	\caption{Results of the random coordinate scaling experiment, which is repeated for 18 times. The dotted line presents the accuracy without coordinate tweaking.}
	\label{story}
\end{figure}

To validate our hypothesis on the different importances of different coordinates, we design a toy experiment. Specifically, we tweak the coordinate feature by multiplying a random scale factor from 0 to 1 to each of the three coordinate ($x, y, z$) and evaluate the accuracy. For simplicity, we use a tiny 5-layer network which is simplified from HCN by removing the convolutions following the transposition operation. The model is trained on four difficult classes of NTU-RGB+D, namely reading, writing, playing with phone/tablet and typing on a keyboard. For each setting of scale factors, we train the model three times using different random seeds, and compute the mean accuracy. The results of coordinate scaling are shown in Figure~\ref{story}. Unsurprisingly, the performance fluctuates greatly as different scale factors are applied to the coordinates and the accuracy without tweaking is not the best. This suggests that the coordinate feature deserves more attention.

\begin{figure}[t]\centering
	\subfigure[CAG]{\label{CAG}\includegraphics[width=0.2\textwidth]{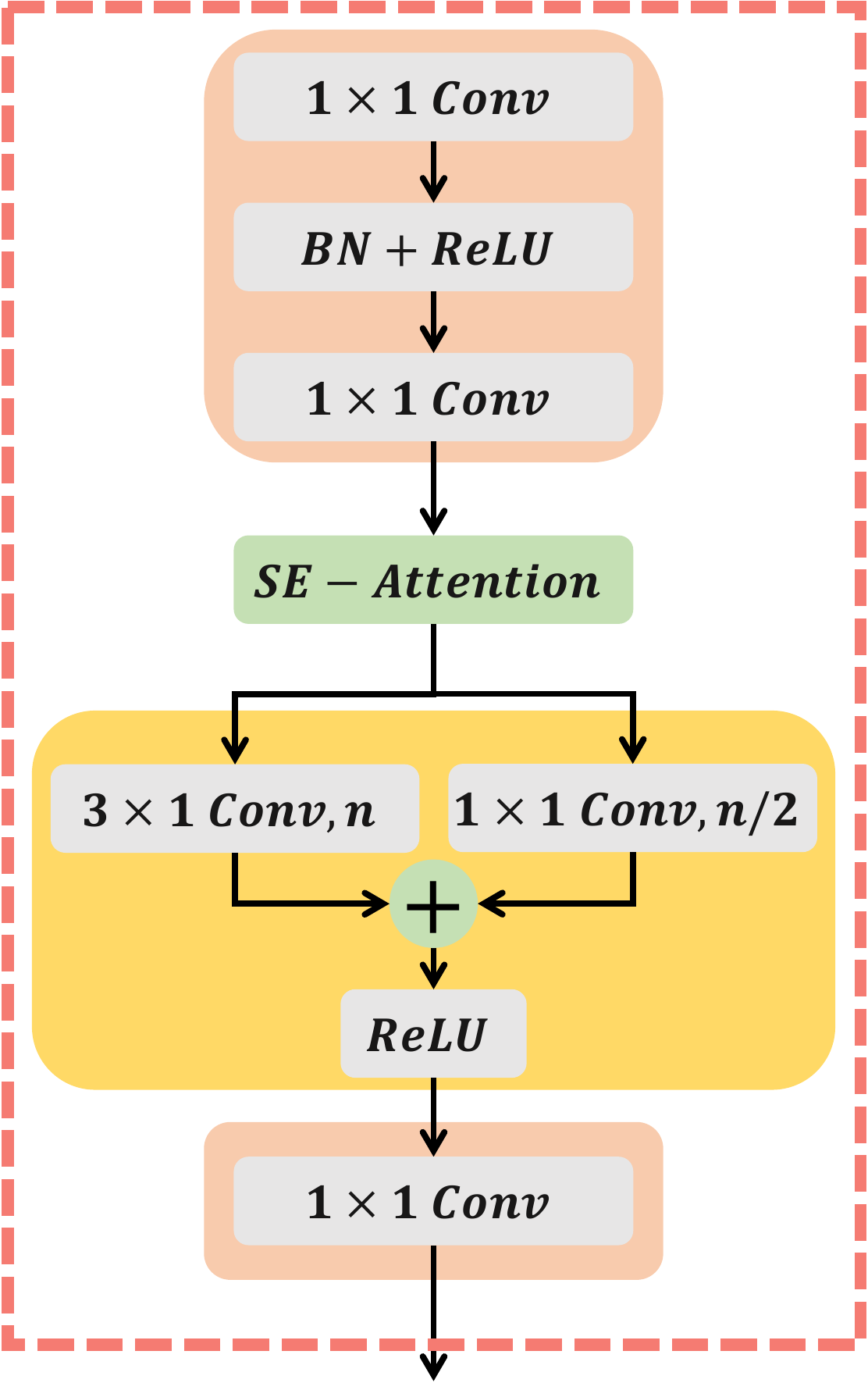}}
	\subfigure[VAG]{\label{VAG}\includegraphics[width=0.2\textwidth]{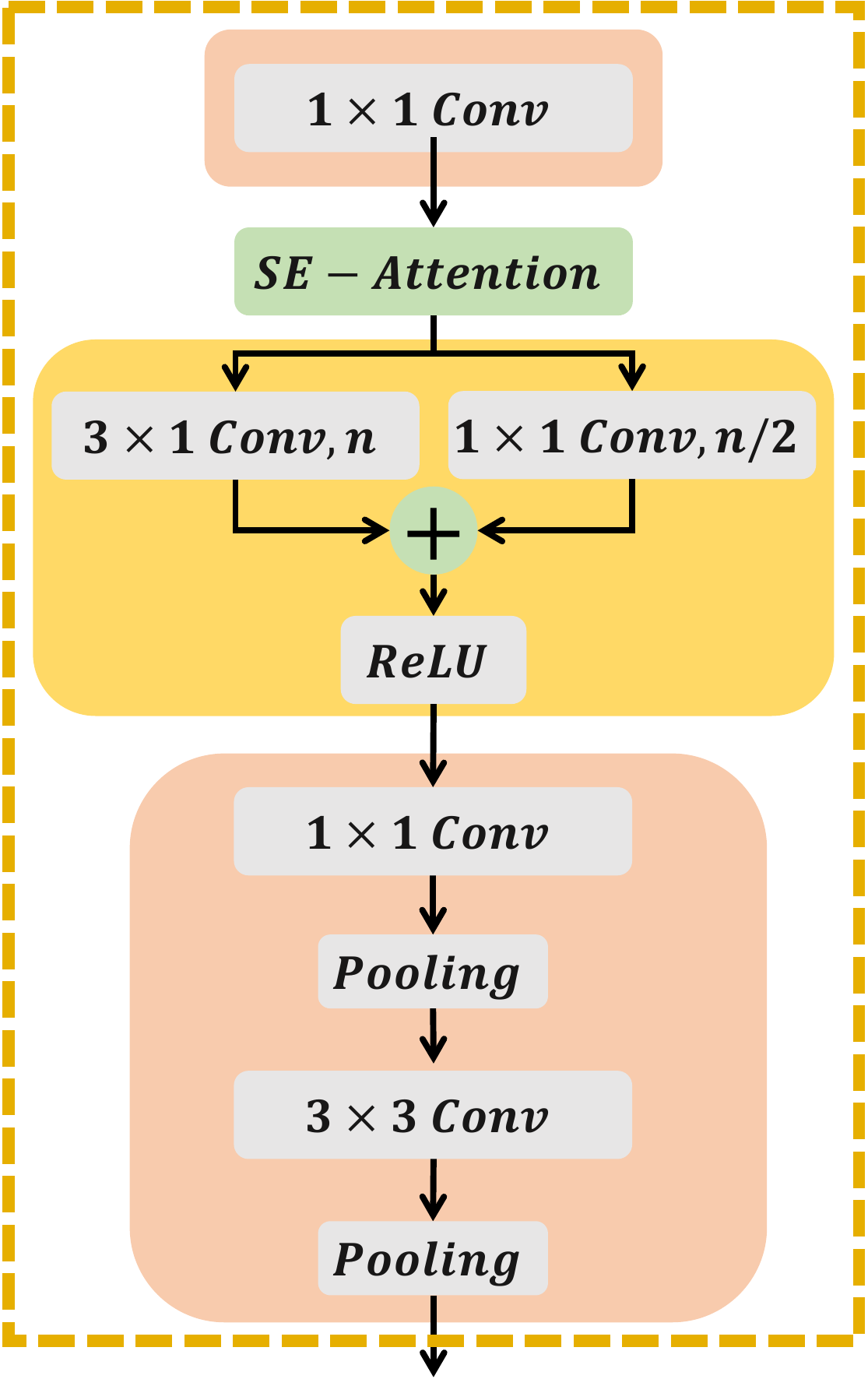}}
    \caption{The detailed architectures of CAG and VAG. The blocks in pink, green and gold denote the feature mapping, channel attention and grouping operations respectively. The grouping operation is implemented as dual coordinate-wise convolution for CAG (a) and dual part-wise convolution for VAG (b).}
	\label{CV}
\end{figure}

The concept of cross-channel feature augmentation is first applied to the coordinate feature, instantiated as the coordinate aware grouping (CAG) module. As shown in Figure~\ref{CAG}, CAG is composed of three main blocks, \ie feature mapping, channel attention and dual coordinate-wise convolution.

\paragraph{Feature Mapping} is mainly constructed with one or a few convolutional layers. It is used to map the coordinate feature into a high-dimensional latent space or reduce the dimension.

\paragraph{Channel Attention} is implemented based on the squeeze-and-excitation attention mechanism~\cite{hu_squeeze-and-excitation_nodate}. It learns channel-wise attention, which essentially enhances discriminative coordinate axes and suppresses unimportant and confusing axes. In other words, this operation helps the model dig out the most useful topology representation implicitly.

\paragraph{Dual Coordinate-wise Convolution} is the key component of CAG. Two parallel grouped convolutions are leveraged to further strengthen the coordinate feature. One is a $3\times 1$ convolution with $n$ groups. The other is a $1\times1$ convolution with $n/2$ groups. The outputs of the two grouped convolutions are fused by element-wise summation. With grouped convolution, the entire high-dimensional feature space is divided into multiple low-dimensional subspaces. Then feature fusion occurs within each subspace separately. Besides, by using different numbers of groups in the two convolutional layers, the feature space is partitioned into subspaces of different granularities. This block effectively enhance the coordinate feature.

The overall process of CAG can be formulated as:
\begin{equation}
\mathbf{Z}=\mathcal{M}_2(\mathcal{G}(\mathcal{A}(\mathcal{M}_1(\mathbf{X})), \dot{\theta})+\mathcal{G}(\mathcal{A}(\mathcal{M}_1(\mathbf{X})), \ddot{\theta}))),
  \label{eq:cag-inpout}
\end{equation}
where $\mathbf{X} \in R^{C\times T \times V}$ and $\mathbf{Z} \in R^{C^{\prime}\times T \times V}$ are the input and output of the CAG module respectively.
$\mathcal{M(\cdot)}$ and $\mathcal{A(\cdot)}$ mean feature mapping and the channel attention individually. $\mathcal{G}(\cdot, \dot{\theta})$ and $\mathcal{G}(\cdot, \ddot{\theta})$ are the two grouped convolutions of the dual coordinate-wise convolution module, which can be expressed as
\begin{equation}
\mathcal{G}(\mathbf{\widetilde{X}}, \theta)=\left[f(\widetilde{X}^{1},\theta^{1}),f(\widetilde{X}^{2},\theta^{2}),\ldots,f(\widetilde{X}^{n},\theta^{n})\right],
\end{equation}
where $\widetilde{\mathbf{X}}=[\widetilde{X}^{1},\widetilde{X}^{2},\ldots,\widetilde{X}^{n}]$ represents the input feature map partitioned into $n$ groups, and $\theta=[\theta^{1},\theta^{2},\ldots,\theta^{n}]$ denotes the weights of the convolutional layer $f$. $[\cdot]$ is the operation of concatenation.

\subsection{Graph Convolution Is a Special Convolution}
Here we illustrate a tight relation between graph convolution and normal convolution. Given a skeleton sequence $\mathbf{X} \in R^{C\times T \times V}$, the graph convolution can be formulated as
\begin{equation}
  \mathbf{Y}=\mathbf{W}\mathbf{X}\mathbf{A},
\end{equation}
where $\mathbf{A} \in R^{V \times V}$ denotes the adjacency matrix representing the skeleton topological information, and $\mathbf{W}\in R^{C^{\prime} \times C }$ is the weighting matrix for feature transformation. For each element in $\mathbf{Y}\in R^{C^{\prime} \times T \times V}$, if we ignore the feature transformation, the core calculation could be shown as
\begin{equation}
  Y_{c,t,v}=\sum_{u=0}^{V-1}X_{c,t,u}A_{u,v}
\label{gcn}
\end{equation}
The essence of GCN is to aggregate the temporal-spatial features over all joints by weighted average guided by the adjacency matrix.

When the joint dimension is transposed as channels, \ie $\mathbf{X}\in R^{V\times T\times C}$, a normal convolution with a kernel size of $P\times Q$ could be formulated as
\begin{equation}
  Y_{v,t,c}=\sum_{u=0}^{V-1}\sum_{p=0}^{P-1}\sum_{q=0}^{Q-1} w_{v,u,p,q}X_{u,t+p-P/2,c+q-Q/2},
\label{cnn}
\end{equation}
where $p$ and $q$ represent the indices of the two-dimensional kernel. When the kernel size is $1\times 1$, Eq.~\eqref{cnn} is simplified to
\begin{equation}
  Y_{v,t,c}=\sum_{u=0}^{V-1} w_{v,u}X_{u,t,c}
\label{conv1x1}
\end{equation}
If the number of output channels is equal to the input channels, and the adjacency matrix is employed as the weights, then the convolution is equivalent to the graph convolution. On the other hand, graph convolution can be considered as a special case of convolution. Notably, with a larger kernel size (\eg $3\times 3$), the convolutional network can perceive not only the topology of joints, but also the temporal feature and coordinate feature jointly. By varying the number of output channels, we are able to expand or reduce the joint dimension, which makes the CNN-based model more flexible and expressive than GCN.

\subsection{Enhancing Joint Feature via VAG}
From the relation between graph convolution and normal convolution, we can conclude that CNN is capable of modeling the topology of joints implicitly. However, we argue that this capability can be further improved. \citeauthor{huang_part-level_2020} split the skeleton into multiple parts explicitly and aggregate features of each part hierarchically. Inspired by this work, we introduce the virtual-part aware grouping module (VAG). It follows the same map-attend-group-map paradigm as CAG does. As shown in Figure~\ref{VAG}, VAG mainly consists of three blocks, namely feature mapping, channel attention and dual part-wise convolution. Note that although the three blocks are analogous to those of CAG, their architectures have been adapted. Here the two grouped convolutions essentially divide the virtual joints into multiple virtual parts, which helps to learn more discriminative joint feature.

\subsection{Ta-CNN Framework}
Figure~\ref{pipline} illustrates the architecture of our Ta-CNN framework. Like HCN, there are also two input streams, \ie the skeleton sequence and the skeleton motion. Accordingly there two sub-networks, both of which are equipped with CAG and VAG. Features of multiple persons are fused by element-wise maximum in a late stage. We tweak the architecture of HCN by adding batch normalization after the first convolutional layer and replacing the first fully-connected layer with a mean operation, which slightly improves the performance with fewer parameters. The modified HCN is adopted as a strong baseline. The detailed configuration of the network will be given in the supplementary material.

\subsection{SkeletonMix}
\begin{figure}[t]
	\centering
	\includegraphics[width=0.45\textwidth]{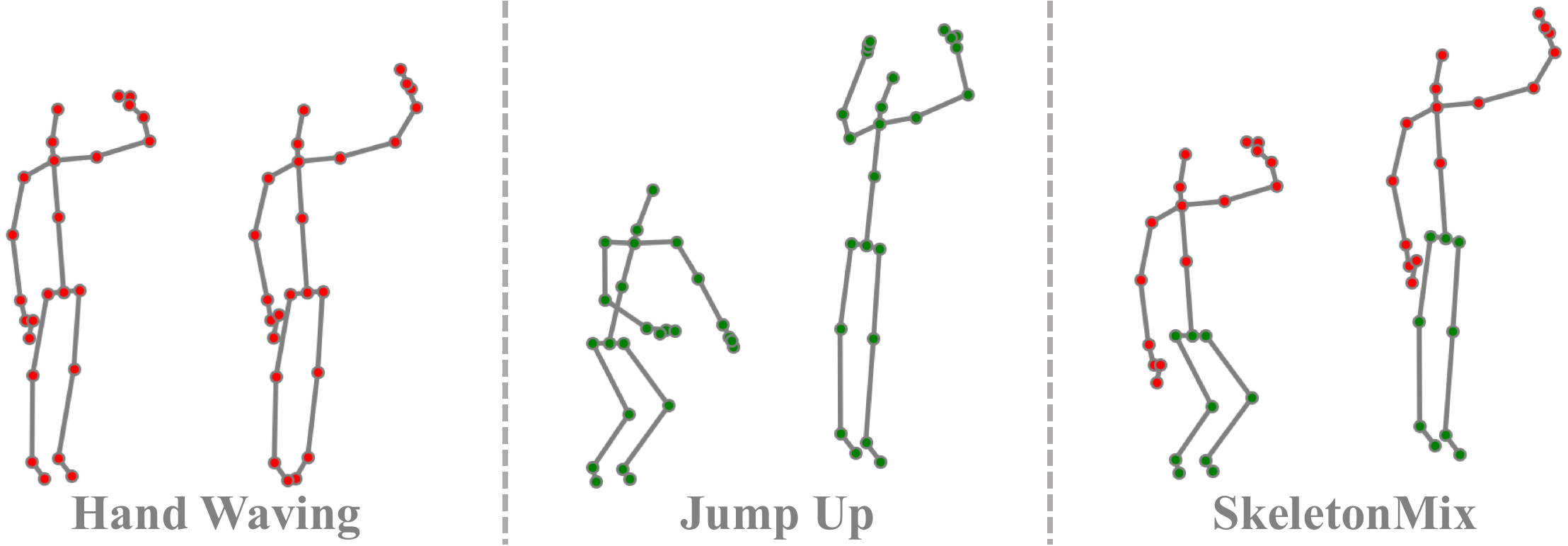} 
	\caption{SkeletonMix augments skeleton data by combining the upper body and the lower body of two persons performing different actions.}
	\label{skeletonmix}
\end{figure}

Mixup~\cite{zhang_mixup_2018} was originally designed for augmentation of image data. It works by mixing two images and their labels linearly by
\begin{equation}
\left\{\begin{array}{l}
\hat{x}=\lambda x_i+(1-\lambda) x_j \\
\hat{y}=\lambda y_i+(1-\lambda) y_j
\end{array}\right.,
\end{equation}
where $(x_i,y_i)$ and $(x_j,y_j)$ are two image-label pairs sampled from the training dataset and $(\hat{x},\hat{y})$ is the output sample. $\lambda\in[0,1]$ is used to control the bias between the two input samples. If we apply the mixup strategy directly to skeleton data, we observe severe deformation to the skeleton topology, leading to poor performance.

Here, we invent a novel mixup strategy specialized for skeleton data. SkeletonMix works by randomly combining the upper body and the lower body of two different persons, which can be formulated as
\begin{equation}
\left\{\begin{array}{l}
\hat{x}=Concat(x_{i,V_u},x_{j,V_l}) \\
\hat{y}=\lambda y_i+(1-\lambda) y_j
\end{array}\right.,
  \label{eq:skeletonmix}
\end{equation}
where $V_u$ and $V_l$ are the joint indices for the upper body and the lower body respectively. Figure~\ref{skeletonmix} shows an example, in which two actions, \ie hand waving and jump up are mixed into a new multi-label action. In our experiments, we randomly select a part of samples in a batch with a certain ratio $\alpha\in[0,1]$ to perform the SkeletonMix augmentation, and the rest retain unchanged.

\section{Experiments}
Our method is evaluated on four widely used datasets, \ie NTU RGB+D, NTU RGB+D 120, Northwestern-UCLA and SBU Kinect Interaction. Extensive ablation studies are conducted to verify the impact of different components of the framework. Finally, performance comparisons with state-of-the-art methods are reported.

\subsection{Datasets}

\paragraph{NTU RGB+D} contains 56,880 samples of 60 classes performed by 40 distinct subjects. There are two recommended benchmarks, \ie cross-subject (X-sub) and cross-view (X-view).

\paragraph{NTU RGB+D 120} is an extension of NTU RGB+D containing 114,480 samples. The numbers of classes and subjects are expanded to 120 and 106 respectively. There are two benchmarks, \ie cross-subject (X-sub) and cross-setup (X-setup).

\paragraph{Northwestern-UCLA (N-UCLA)} is captured by using three Kinect cameras. It contains 1494 samples of 10 classes. The same evaluation protocol in~\citeauthor{wang_cross-view_2014} is adopted.

\paragraph{SBU Kinect Interaction (SBU)} depicts human interaction captured by Kinect. It contains 282 sequences covering 8 action classes. Subject-independent 5-fold cross-validation is performed following the same evaluation protocol as previous works~\cite{yun_two-person_2012}.

\subsection{Implementation Details}\label{ID}
The proposed model is implemented in PyTorch~\cite{paszke_automatic_2017}. The model is trained for 800 epochs with the Adam optimizer. The learning rate is set to 0.001 initially and decayed by a factor of 0.1 at 650, 730 and 770 epochs for all datasets. The weight decay is set to 0.0002 for SBU and 0.0001 for the rest datasets. The batch sizes for NTU RGB+D, NTU RGB+D 120, N-UCLA and SBU are 64, 64, 16 and 8 respectively. As for data pre-processing method, we follow previous works~\cite{shi_two-stream_2019,cheng_skeleton-based_2020,li_co-occurrence_2018}. For SBU, a warmup strategy is used during the first 30 epochs. The $\lambda$ in Eq.~\eqref{eq:skeletonmix} is set to 0.6, and the mixing ratio $\alpha$ for SkeletonMix is set to $1/16$.

\subsection{Ablation Studies}

\begin{table}[t]
	\centering
  \small
  \setlength{\tabcolsep}{5pt}
	\begin{tabular}{l|c|c|c}
		\toprule
        Method & Param. (M) & GFLOPs & Acc. (\%)\\
		\midrule
        HCN~\cite{li_co-occurrence_2018}   &  0.78  &  0.06 & 86.5 \\
		Baseline & 0.54 & 0.11 & 87.4$(\uparrow 0.9)$  \\
		+VAG (6) & 0.53 & 0.09   & 87.8 $(\uparrow 1.3)$ \\
		+VAG (10) &  0.53 &  0.09 & 87.8 $(\uparrow 1.3)$ \\
		+VAG (10)+CAG (10) &  0.53  &  0.08 & 88.7 $(\uparrow 2.2)$ \\
		+VAG (6)+CAG (10) & 0.53 & 0.08  & 88.8 $(\uparrow 2.3)$ \\
		\bottomrule
	\end{tabular}
	\caption{Accuracy gains of VAG and CAG on NTU RGB+D.}
	\label{CAGVAG}%
\end{table}

\begin{table}[t]
	\centering
    \small
	\begin{tabular}{l|c|c|c}
		\toprule
        Method & Param. (M) & GFLOPs & Acc. (\%)\\
		\midrule
		Baseline &  0.54  & 0.11  & 87.4 \\
		+CA   &  0.56  &  0.11 & 87.9 $(\uparrow 0.5)$ \\
		+CA+FM & 0.54  &  0.12 & 88.3 $(\uparrow 0.9)$\\
		+CA+FM+SGC &  0.53   & 0.08  & 88.1 $(\uparrow 0.7)$ \\
		+CA+FM+DGC &  0.53  & 0.08  & 88.8 $(\uparrow 1.4)$ \\
		\bottomrule
	\end{tabular}%
    \caption{Accuracy gains by channel attention (CA), feature mapping (FM) and dual grouped convolution (DGC)  on NTU RGB+D. SGC represents single grouped convolution.}
	\label{cm}%
\end{table}%

\subsubsection{Effectiveness of CAG and VAG.}
Table~\ref{CAGVAG} shows the performance gains brought by CAG and VAG with the cross-subject benchmark on the NTU RGB+D dataset. We first evaluate the strong baseline model which is derived from HCN. It improves the accuracy of HCN by 0.9\% with fewer parameters. Starting from the strong baseline, VAG improves the accuracy slightly with GFLOPs reduced from 0.11 to 0.9. VAG with $n=6$ and $n=10$ achieve similar accuracy. CAG further boosts the accuracy by 1\% without increasing the parameters and GFLOPs. In total we achieve an accuracy of 88.8\%, which improves the accuracy reported in HCN by 2.3\% with a more efficient model.

\subsubsection{Impact of the Blocks in CAG \& VAG.}
Both CAG and VAG are composed of the feature mapping (FM), channel attention (CA) and dual grouped convolution (DGC) blocks. Table~\ref{cm} shows their individual contribution to the final performance. CA, FM and DGC improve the accuracy by 0.5\%, 0.4\% and 0.5\% respectively. It is worth noting that single grouped convolution (SGC) slightly harms the performance as the GFLOPs gets reduced. This indicates that grouped convolution alone does not suffice to enhance the topology feature, and the design of two convolutions with different numbers of groups is necessary.

\subsubsection{Visualization of Channel Attention.}
\begin{figure}[t]\centering
	\subfigure[Coordinate attention]{\label{cse}\includegraphics[height=0.3\textwidth]{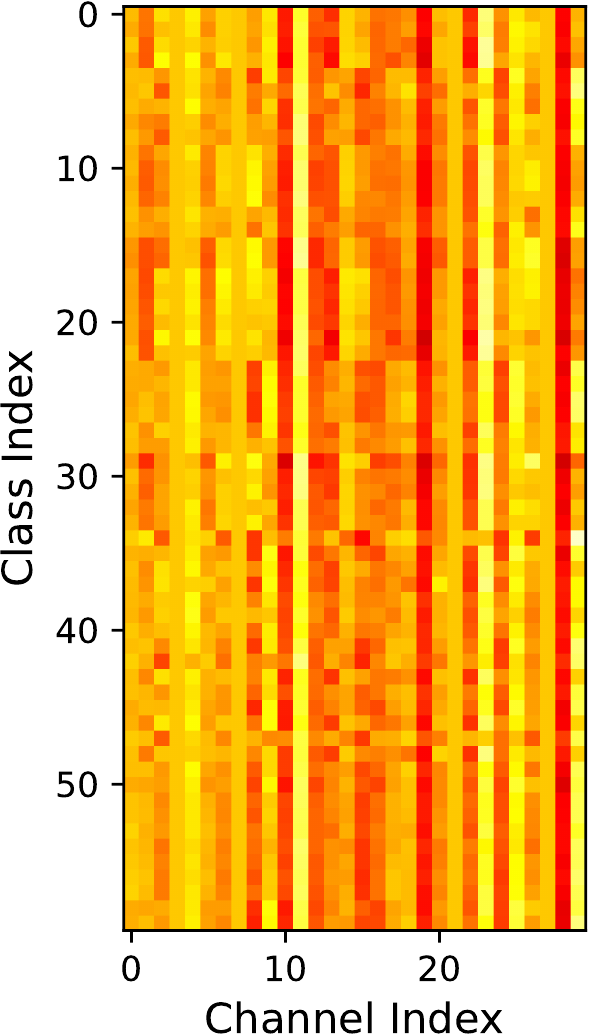}}
	\subfigure[Joint attention]{\label{vse}\includegraphics[height=0.3\textwidth]{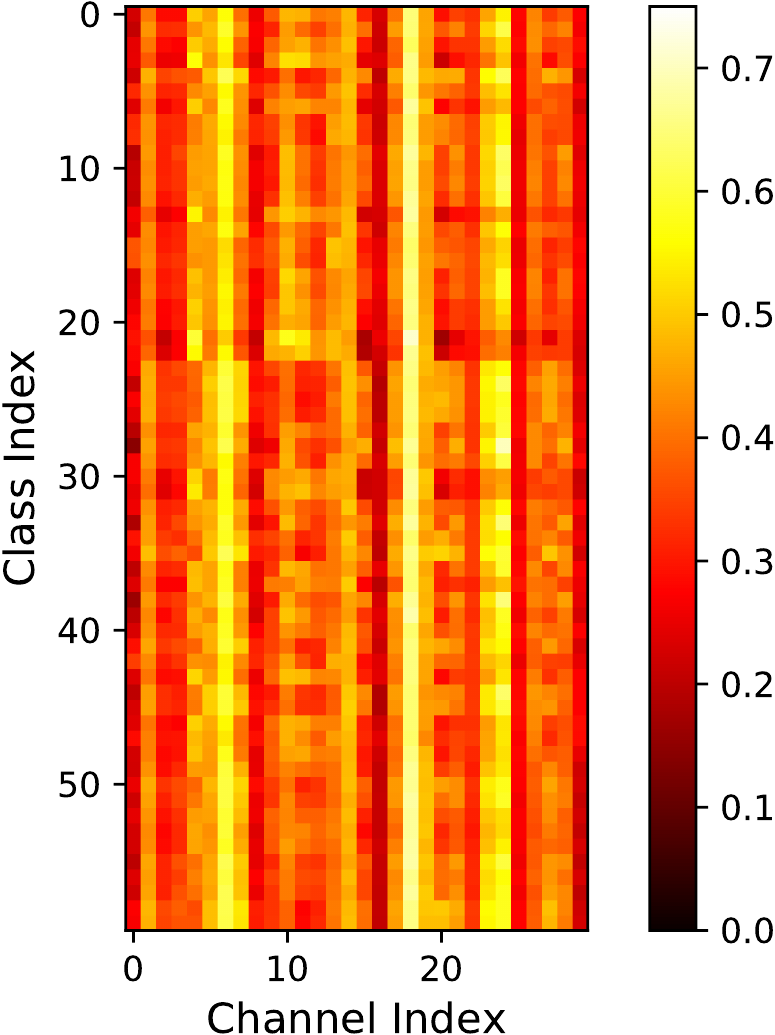}}
    \caption{Visualization of the learned (a) coordinate attention in CAG and (b) joint attention in VAG.}
	\label{se}
\end{figure}

Figure~\ref{se} visualizes the learned channel attention on the NTU RGB+D dataset. The two heat maps show the average attention response for each channel and each class in the test dataset in CAG and VAG of the skeleton sequence branch. We can see for some channels, the attention values are highly correlated between different classes. While for others, the response varies for different classes. This confirms our analysis that the dimensions of coordinate and joint features are not equally important for recognition of the action.

\subsubsection{Effectiveness of SkeletonMix.}
\begin{table}[t]
	\centering
	\begin{tabular}{l|c|c|c|c}
		\toprule
		\multicolumn{1}{l|}{\multirow{2}[4]{*}{Method}} & \multicolumn{2}{c|}{w/o SkeletonMix} & \multicolumn{2}{c}{w/ SkeletonMix} \\
		\cmidrule{2-5}          & \multicolumn{1}{c|}{X-sub} & \multicolumn{1}{c|}{X-setup} & \multicolumn{1}{c|}{X-sub} & \multicolumn{1}{c}{X-setup} \\
		\midrule
        HCN~\shortcite{li_co-occurrence_2018} & 76.5  & 75.1  & 76.5  &  78.2 \\
        2s-AGCN~\shortcite{shi_two-stream_2019} & 76.4  & 78.4  & 76.9  & 80.0  \\
		\bottomrule
	\end{tabular}%
	\caption{Performance of SkeletonMix combined with HCN and 2s-AGCN on NTU RGB+D 120.}
	\label{PM2}%
\end{table}%

\begin{table}[t]
	\centering
	\begin{tabular}{l|c|c|c|c}
		\toprule
		\multicolumn{1}{l|}{\multirow{2}[4]{*}{Modality}} & \multicolumn{2}{c|}{w/o SkeletonMix} & \multicolumn{2}{c}{w/ SkeletonMix} \\
		\cmidrule{2-5}          & \multicolumn{1}{c|}{X-sub} & \multicolumn{1}{c|}{X-setup} & \multicolumn{1}{c|}{X-sub} & \multicolumn{1}{c}{X-setup} \\
		\midrule
		Joint & 82.1  & 83.5  & 82.4  &  84.0 \\
		Bone  & 82.3  & 84.0  & 82.6  & 84.4  \\
		Joint Motion & 77.2  & 78.7  & 77.5  &  79.0 \\
		Bone Motion  & 76.9  & 78.6  & 77.1  & 78.9  \\
		\bottomrule
	\end{tabular}%
	\caption{Performance of SkeletonMix when trained using various data modalities on NTU RGB+D 120.}
	\label{PM1}%
\end{table}%

\begin{table}[t]
	\centering
    \small
  \setlength{\tabcolsep}{4pt}
	\begin{tabular}{c|c|c|c|c}
		\toprule
        Method & \multicolumn{1}{c|}{Baseline} & \multicolumn{1}{c|}{Mixup} & SMix ($\alpha=1$) & SMix ($\alpha=1/16$)\\
		\midrule
		Acc. (\%) & 82.1  & 79.8  & 75.7  & 82.4 \\
		\bottomrule
	\end{tabular}%
    \caption{Comparison of the original mixup and SkeletonMix (SMix) with two mixing ratios ($\alpha$) on NTU RGB+D 120.}
	\label{tab:ablation-sm}%
\end{table}%

\begin{table}[!t]
	\small
  \setlength{\tabcolsep}{4pt}
	\centering
	\begin{tabular}{c|l|c|c|c}
		\toprule
        Type & \multicolumn{1}{|l}{Method} & \multicolumn{1}{|c}{Param. (M)} & \multicolumn{1}{|c|}{GFLOPs} & \multicolumn{1}{c}{Acc. (\%)} \\
		\midrule
		\multirow{6}[0]{*}{CNN} & HCN \shortcite{li_co-occurrence_2018} & 0.78 & 0.06     & 86.5 \\
		&SResNet101 \shortcite{zhang_view_2019} & 42.62 & -     & 87.8 \\
		&SResNet152 \shortcite{zhang_view_2019} & 58.27 & -     & 88.2 \\
		&VResNet10 \shortcite{zhang_view_2019} & 5.39  & -     & 83.0 \\
		&VResNet18 \shortcite{zhang_view_2019} & 11.66 & -     & 87.7 \\
		&VResNet50 \shortcite{zhang_view_2019} & 24.09 & -     & 88.7 \\
		\midrule
		\midrule
		\multirow{7}[0]{*}{GCN} &Shift-GCN \shortcite{cheng_skeleton-based_2020} & 0.69  & 2.50   & 87.8 \\
		&Shift-GCN++ \shortcite{cheng_extremely_2021} & 0.45  & 0.40   & 87.9 \\
		&SGN-5 \shortcite{zhang_semantics-guided_2020}   & 0.69 & 0.80   & 89.0 \\
		&SGN-1 \shortcite{zhang_semantics-guided_2020}   & 0.69 & 0.16   & 88.4 \\
		&AdaSGN \shortcite{shi_adasgn_2021} & 2.05   & 0.07  & 89.1 \\
		&AdaSGN-Fuse \shortcite{shi_adasgn_2021} & 2.05   & 0.32  & 89.3 \\
		&MS-G3D \shortcite{liu_disentangling_2020} & 3.20   & 24.4  & 89.4 \\
		\midrule
		\multirow{2}[0]{*}{Ours} & Ta-CNN  & 0.53   & 0.08  & 88.8 \\
		& Ta-CNN+  & 1.06     & 0.16  & 89.6 \\
		\bottomrule
	\end{tabular}%
	\caption{The effectiveness and efficiency of Ta-CNN compared with the state-of-the-art methods on NTU RGB+D. SResNet and VResNet mean S-trans+ResNet and VA-ResNet respectively.}
	\label{eff}%
\end{table}%

To validate the proposed SkeletonMix strategy, we conduct experiments on the more challenging NTU RGB+D 120 dataset using both cross-subject and cross-setup benchmarks. Apart from the proposed Ta-CNN, we also apply it to two existing methods, \ie HCN and 2s-AGCN. As compared in Table~\ref{PM2}, both HCN and 2s-AGCN benefit from the strategy a lot, especially in the cross-setup setting. This proves the generalizability of SkeletonMix. As shown in Table~\ref{PM1}, for Ta-CNN, no matter what data modality is used as input, it can bring considerable improvement. Note that as compared in Table~\ref{tab:ablation-sm}, applying the original mixup strategy directly to skeleton data leads to performance degradation. And the mixing ratio ($\alpha$) of SkeletonMix is also important.

\subsubsection{Efficiency of Ta-CNN.}
As a pure CNN-based method, we compare the proposed Ta-CNN with previous CNN-based and GCN-based methods in terms of the number of parameters and GFLOPs. As summarized in Table~\ref{eff}, we achieve state-of-the-art performance while balancing the efficiency. Ta-CNN+ is an ensemble of two Ta-CNNs with different $n$ (6 and 10) in VAG. Although the computational complexity is doubled, it is still competitive considering the improved accuracy. This characteristic makes it very suitable for deployment, especially on edge devices where computational and memory resources are limited.

\subsection{Comparison with the State-of-the-arts}

\begin{table}[t]
	\small
	\centering
	\begin{tabular}{c|l|c}
		\toprule
		Dataset&\multicolumn{1}{|l|}{Method} & \multicolumn{1}{c}{Acc. (\%)} \\
		\midrule
		\multirow{14}{*}{N-UCLA} &Lie Group~\shortcite{vemulapalli_human_2014} & 74.2 \\
		& Actionlet ensemble~\shortcite{wang_learning_2014} & 76.0 \\
		& HBRNN-L~\shortcite{yong_du_hierarchical_2015} & 78.5 \\
		& Ensemble TS-LSTM~\shortcite{lee_ensemble_2017} & 89.2 \\
		& VA-CNN (aug.)~\shortcite{zhang_view_2019} & 90.7 \\
		& VA-RNN (aug.)~\shortcite{zhang_view_2019} & 93.8 \\
		& VA-fusion (aug.)~\shortcite{zhang_view_2019} & 95.3 \\
		& AGC-LSTM~\shortcite{si_attention_2019} & 93.3 \\
		& Shift-GCN~\shortcite{cheng_skeleton-based_2020} & 94.6 \\
		& Shift-GCN++~\shortcite{cheng_extremely_2021} & 95.0 \\
		& DC-GCN+ADG~\shortcite{vedaldi_decoupling_2020} & 95.3 \\
		& CTR-GCN~\shortcite{chen_channel-wise_2021} & 96.5 \\
		\cmidrule{2-3}          &Ta-CNN  & 96.1 \\
		&Ta-CNN+  & 97.2 \\
		\midrule
		\midrule
		\multirow{12}[0]{*}{SBU}&{HBRNN-L~\shortcite{yong_du_hierarchical_2015}} & 80.4 \\
		&Co-occurrence RNN~\shortcite{zhu_co-occurrence_2016} & 90.4 \\
		&STA-LSTM~\shortcite{song_end--end_2017} & 91.5 \\
		&ST-LSTM + Trust Gate~\shortcite{liu_spatio-temporal_2016} & 93.3 \\
		&GCA-LSTM~\shortcite{liu_global_2017} & 94.1 \\
		&Clips+CNN+MTLN~\shortcite{ke_new_2017} & 93.6 \\
		&VA-RNN (aug.)~\shortcite{zhang_view_2019} & 97.5 \\
		&VA-CNN (aug.)~\shortcite{zhang_view_2019} & 95.7 \\
		&VA-fusion (aug.)~\shortcite{zhang_view_2019}& 98.3 \\
		&HCN~\shortcite{li_co-occurrence_2018} & 98.6 \\
		\cmidrule{2-3}          &Ta-CNN  & 98.5 \\
		&Ta-CNN+  & 98.9 \\
		\bottomrule
	\end{tabular}%
	\caption{Comparisons of the top-1 accuracy ($\%$) with the state-of-the-art methods on N-UCLA and SBU.}
	\label{N-UCLA}%
\end{table}%

\begin{table}[t]
	\small
	\centering
	
	\begin{tabular}{c|l|c|c}
		\toprule
		Type & \multicolumn{1}{l|}{Method} & \multicolumn{1}{c|}{X-sub} & \multicolumn{1}{c}{X-view} \\
		\midrule
		\multirow{7}{*}{Heavy GCN}&2s-AGCN~\shortcite{shi_two-stream_2019} & 88.5  & 95.1  \\
		&PL-GCN~\shortcite{huang_part-level_2020} & 89.2  & 95.0  \\
		&AGC-LSTM~\shortcite{si_attention_2019} & 89.2  & 95.0  \\
		&Dynamic GCN~\shortcite{ye_dynamic_2020} & 91.5  & 96.0  \\
		&MST-GCN~\shortcite{chen_multi-scale_2021} & 91.5  & 96.6  \\
		&MS-G3D~\shortcite{liu_disentangling_2020} & 91.5  & 96.2  \\
		&CTR-GCN~\shortcite{chen_channel-wise_2021} & 92.4  & 96.8  \\
		\midrule\midrule
		\multirow{4}{*}{Light GCN}&SGN~\shortcite{zhang_semantics-guided_2020} & 89.0  & 94.5  \\
		&Shift-GCN~\shortcite{cheng_skeleton-based_2020} & 90.7  & 96.5  \\
		&Shift-GCN++~\shortcite{cheng_extremely_2021} & 90.5  & 96.3  \\
		&3s-AdaSGN~\shortcite{shi_adasgn_2021} & 90.5  & 95.3  \\
		\midrule\midrule
		\multirow{6}{*}{CNN}&Res-TCN~\shortcite{kim_interpretable_2017}& 74.3  & 83.1  \\
		&Two-stream CNN~\shortcite{li_skeleton-based_2017} & 83.2  & 89.3  \\
		&HCN~\shortcite{li_co-occurrence_2018} & 86.5  & 91.1  \\
		&VA-CNN (aug.)~\shortcite{zhang_view_2019}& 88.7  & 94.3  \\
		&VA-fusion (aug.)~\shortcite{zhang_view_2019} & 89.4  & 95.0  \\
		\midrule
		\multirow{2}{*}{Ours}&Ta-CNN  & 90.4  & 94.8  \\
		&Ta-CNN+  & 90.7  & 95.1  \\
		\bottomrule
	\end{tabular}%
	\caption{Comparisons of the top-1 accuracy ($\%$) with the state-of-the-art methods on NTU RGB+D.}
	\label{tab:ntu60}%
\end{table}%
	
	\begin{table}[!t]
	\centering
      \small
	\begin{tabular}{c|l|c|c}
		\toprule
		Type &\multicolumn{1}{l|}{Method} & \multicolumn{1}{c|}{X-sub} & \multicolumn{1}{c}{X-setup} \\
		\midrule
		\multirow{5}{*}{Heavy GCN}&2s-AGCN~\shortcite{shi_two-stream_2019} & 82.5  & 84.2  \\
		&Dynamic GCN~\shortcite{ye_dynamic_2020} & 87.3  & 88.6  \\
		&MST-GCN~\shortcite{chen_multi-scale_2021} & 87.5  & 88.8  \\
		&MS-G3D~\shortcite{liu_disentangling_2020} & 86.9  & 88.4  \\
		&CTR-GCN~\shortcite{chen_channel-wise_2021} & 88.9  & 90.6  \\
		\midrule\midrule
		\multirow{4}{*}{Light GCN}&SGN~\shortcite{zhang_semantics-guided_2020} & 79.2  & 81.5  \\
		&Shift-GCN~\shortcite{cheng_skeleton-based_2020} & 85.9  & 87.6  \\
		&Shift-GCN++~\shortcite{cheng_extremely_2021} & 85.6  & 87.2  \\
		&3s-AdaSGN~\shortcite{shi_adasgn_2021} & 85.9  & 86.8  \\
		\midrule
		\multirow{2}{*}{Ours}&Ta-CNN  &  85.4  & 86.8  \\
		&Ta-CNN+  &  85.7  & 87.3  \\
		\bottomrule
	\end{tabular}%
	\caption{Comparisons of the top-1 accuracy ($\%$) with the state-of-the-art methods on NTU RGB+D 120.}
	\label{tab:ntu120}%
\end{table}%

Here, we compare our final performance with existing methods on the aforementioned four datasets. On the N-UCLA and CBU datasets, we obtain an accuracy of 97.2\% and 98.9\% respectively (see Table~\ref{N-UCLA}), setting the new state-of-the-art. On NTU RGB+D (Table~\ref{tab:ntu60}) and NTU RGB+D 120 (Table~\ref{tab:ntu120}), we outperform all existing CNN-based methods by a large margin with much fewer parameters and calculation amounts. Compared with the well developed GCN-based methods, we achieve comparable performance with the superiority in terms of model size.

\section{Conclusion}
This paper proposes a novel pure CNN model for skeleton-based action recognition. By rethinking the weakness of CNN in encoding the irregular skeleton topology, we are committed to enhance the topology feature. With a carefully designed cross-channel feature augmentation module and a mixup strategy specialized for skeleton data, we achieve state-of-the-art performance with a tiny model. In addition, we theoretically prove that graph convolution is essentially a special case of normal convolution when the joint dimension is treated as channels. The conclusion is consistent with our model design. We argue that the potential of CNN for modeling of irregular graph data beyond skeleton data has not been fully exploited and deserves more attention. The excellent performance and tiny model size make our method well suitable for real-world deployment.

{\small
\bibliography{main}

\begin{thebibliography}{43}
\providecommand{\natexlab}[1]{#1}

\bibitem[{Chen et~al.(2021{\natexlab{a}})Chen, Zhang, Yuan, Li, Deng, and
  Hu}]{chen_channel-wise_2021}
Chen, Y.; Zhang, Z.; Yuan, C.; Li, B.; Deng, Y.; and Hu, W. 2021{\natexlab{a}}.
\newblock Channel-wise {Topology} {Refinement} {Graph} {Convolution} for
  {Skeleton}-{Based} {Action} {Recognition}.
\newblock \emph{arXiv:2107.12213 [cs]}.
\newblock ArXiv: 2107.12213.

\bibitem[{Chen et~al.(2021{\natexlab{b}})Chen, Li, Yang, Li, and
  Liu}]{chen_multi-scale_2021}
Chen, Z.; Li, S.; Yang, B.; Li, Q.; and Liu, H. 2021{\natexlab{b}}.
\newblock Multi-{Scale} {Spatial} {Temporal} {Graph} {Convolutional} {Network}
  for {Skeleton}-{Based} {Action} {Recognition}.
\newblock \emph{Proceedings of the AAAI Conference on Artificial Intelligence},
  35(2): 1113--1122.
\newblock Number: 2.

\bibitem[{Cheng et~al.(2020{\natexlab{a}})Cheng, Zhang, Cao, Shi, Cheng, and
  Lu}]{vedaldi_decoupling_2020}
Cheng, K.; Zhang, Y.; Cao, C.; Shi, L.; Cheng, J.; and Lu, H.
  2020{\natexlab{a}}.
\newblock Decoupling {GCN} with {DropGraph} {Module} for {Skeleton}-{Based}
  {Action} {Recognition}.
\newblock In Vedaldi, A.; Bischof, H.; Brox, T.; and Frahm, J.-M., eds.,
  \emph{Computer {Vision} – {ECCV} 2020}, volume 12369, 536--553. Cham:
  Springer International Publishing.
\newblock ISBN 978-3-030-58585-3 978-3-030-58586-0.
\newblock Series Title: Lecture Notes in Computer Science.

\bibitem[{Cheng et~al.(2020{\natexlab{b}})Cheng, Zhang, He, Chen, Cheng, and
  Lu}]{cheng_skeleton-based_2020}
Cheng, K.; Zhang, Y.; He, X.; Chen, W.; Cheng, J.; and Lu, H.
  2020{\natexlab{b}}.
\newblock Skeleton-{Based} {Action} {Recognition} {With} {Shift} {Graph}
  {Convolutional} {Network}.
\newblock In \emph{2020 {IEEE}/{CVF} {Conference} on {Computer} {Vision} and
  {Pattern} {Recognition} ({CVPR})}, 180--189. Seattle, WA, USA: IEEE.
\newblock ISBN 978-1-72817-168-5.

\bibitem[{Cheng et~al.(2021)Cheng, Zhang, He, Cheng, and
  Lu}]{cheng_extremely_2021}
Cheng, K.; Zhang, Y.; He, X.; Cheng, J.; and Lu, H. 2021.
\newblock Extremely {Lightweight} {Skeleton}-{Based} {Action} {Recognition}
  {With} {ShiftGCN}++.
\newblock \emph{IEEE Transactions on Image Processing}, 30: 7333--7348.
\newblock Conference Name: IEEE Transactions on Image Processing.

\bibitem[{Cho et~al.(2014)Cho, van Merrienboer, Gulcehre, Bahdanau, Bougares,
  Schwenk, and Bengio}]{cho_learning_2014}
Cho, K.; van Merrienboer, B.; Gulcehre, C.; Bahdanau, D.; Bougares, F.;
  Schwenk, H.; and Bengio, Y. 2014.
\newblock Learning {Phrase} {Representations} using {RNN} {Encoder}-{Decoder}
  for {Statistical} {Machine} {Translation}.
\newblock \emph{arXiv:1406.1078 [cs, stat]}.
\newblock ArXiv: 1406.1078.

\bibitem[{Du, Fu, and Wang(2015)}]{du_skeleton_2015}
Du, Y.; Fu, Y.; and Wang, L. 2015.
\newblock Skeleton based action recognition with convolutional neural network.
\newblock In \emph{2015 3rd {IAPR} {Asian} {Conference} on {Pattern}
  {Recognition} ({ACPR})}, 579--583.
\newblock ISSN: 2327-0985.

\bibitem[{Hu, Shen, and Sun(2017)}]{hu_squeeze-and-excitation_nodate}
Hu, J.; Shen, L.; and Sun, G. 2017.
\newblock Squeeze-and-{Excitation} {Networks}.
\newblock 10.

\bibitem[{Huang et~al.(2020)Huang, Huang, Ouyang, and
  Wang}]{huang_part-level_2020}
Huang, L.; Huang, Y.; Ouyang, W.; and Wang, L. 2020.
\newblock Part-{Level} {Graph} {Convolutional} {Network} for {Skeleton}-{Based}
  {Action} {Recognition}.
\newblock \emph{AAAI}, 34(07): 11045--11052.
\newblock Number: 07.

\bibitem[{Ke et~al.(2017)Ke, Bennamoun, An, Sohel, and Boussaid}]{ke_new_2017}
Ke, Q.; Bennamoun, M.; An, S.; Sohel, F.; and Boussaid, F. 2017.
\newblock A {New} {Representation} of {Skeleton} {Sequences} for {3D} {Action}
  {Recognition}.
\newblock 10.

\bibitem[{Kim and Reiter(2017)}]{kim_interpretable_2017}
Kim, T.~S.; and Reiter, A. 2017.
\newblock Interpretable {3D} {Human} {Action} {Analysis} with {Temporal}
  {Convolutional} {Networks}.
\newblock In \emph{2017 {IEEE} {Conference} on {Computer} {Vision} and
  {Pattern} {Recognition} {Workshops} ({CVPRW})}, 1623--1631. Honolulu, HI,
  USA: IEEE.
\newblock ISBN 978-1-5386-0733-6.

\bibitem[{Lee et~al.(2017)Lee, Kim, Kang, and Lee}]{lee_ensemble_2017}
Lee, I.; Kim, D.; Kang, S.; and Lee, S. 2017.
\newblock Ensemble {Deep} {Learning} for {Skeleton}-{Based} {Action}
  {Recognition} {Using} {Temporal} {Sliding} {LSTM} {Networks}.
\newblock In \emph{2017 {IEEE} {International} {Conference} on {Computer}
  {Vision} ({ICCV})}, 1012--1020. Venice: IEEE.
\newblock ISBN 978-1-5386-1032-9.

\bibitem[{Li et~al.(2017{\natexlab{a}})Li, Zhong, Xie, and
  Pu}]{li_skeleton-based_2017}
Li, C.; Zhong, Q.; Xie, D.; and Pu, S. 2017{\natexlab{a}}.
\newblock Skeleton-based {Action} {Recognition} with {Convolutional} {Neural}
  {Networks}.
\newblock \emph{arXiv:1704.07595 [cs]}.
\newblock ArXiv: 1704.07595.

\bibitem[{Li et~al.(2018)Li, Zhong, Xie, and Pu}]{li_co-occurrence_2018}
Li, C.; Zhong, Q.; Xie, D.; and Pu, S. 2018.
\newblock Co-occurrence {Feature} {Learning} from {Skeleton} {Data} for
  {Action} {Recognition} and {Detection} with {Hierarchical} {Aggregation}.
\newblock \emph{arXiv:1804.06055 [cs]}.
\newblock HCN.

\bibitem[{Li et~al.(2017{\natexlab{b}})Li, Wen, Chang, Lim, and
  Lyu}]{li_adaptive_2017}
Li, W.; Wen, L.; Chang, M.-C.; Lim, S.~N.; and Lyu, S. 2017{\natexlab{b}}.
\newblock Adaptive {RNN} {Tree} for {Large}-{Scale} {Human} {Action}
  {Recognition}.
\newblock In \emph{2017 {IEEE} {International} {Conference} on {Computer}
  {Vision} ({ICCV})}, 1453--1461. Venice: IEEE.
\newblock ISBN 978-1-5386-1032-9.

\bibitem[{Liu et~al.(2016)Liu, Shahroudy, Xu, and
  Wang}]{liu_spatio-temporal_2016}
Liu, J.; Shahroudy, A.; Xu, D.; and Wang, G. 2016.
\newblock Spatio-{Temporal} {LSTM} with {Trust} {Gates} for {3D} {Human}
  {Action} {Recognition}.
\newblock In Leibe, B.; Matas, J.; Sebe, N.; and Welling, M., eds.,
  \emph{Computer {Vision} – {ECCV} 2016}, Lecture {Notes} in {Computer}
  {Science}, 816--833. Cham: Springer International Publishing.
\newblock ISBN 978-3-319-46487-9.

\bibitem[{Liu et~al.(2017)Liu, Wang, Hu, Duan, and Kot}]{liu_global_2017}
Liu, J.; Wang, G.; Hu, P.; Duan, L.-Y.; and Kot, A.~C. 2017.
\newblock Global {Context}-{Aware} {Attention} {LSTM} {Networks} for {3D}
  {Action} {Recognition}.
\newblock In \emph{2017 {IEEE} {Conference} on {Computer} {Vision} and
  {Pattern} {Recognition} ({CVPR})}, 3671--3680. Honolulu, HI: IEEE.
\newblock ISBN 978-1-5386-0457-1.

\bibitem[{Liu, Liu, and Chen(2017)}]{liu_enhanced_2017}
Liu, M.; Liu, H.; and Chen, C. 2017.
\newblock Enhanced skeleton visualization for view invariant human action
  recognition.
\newblock \emph{Pattern Recognition}, 68: 346--362.

\bibitem[{Liu et~al.(2020)Liu, Zhang, Chen, Wang, and
  Ouyang}]{liu_disentangling_2020}
Liu, Z.; Zhang, H.; Chen, Z.; Wang, Z.; and Ouyang, W. 2020.
\newblock Disentangling and {Unifying} {Graph} {Convolutions} for
  {Skeleton}-{Based} {Action} {Recognition}.
\newblock 143--152.

\bibitem[{Paszke et~al.(2017)Paszke, Gross, Chintala, Chanan, Yang, DeVito,
  Lin, Desmaison, Antiga, and Lerer}]{paszke_automatic_2017}
Paszke, A.; Gross, S.; Chintala, S.; Chanan, G.; Yang, E.; DeVito, Z.; Lin, Z.;
  Desmaison, A.; Antiga, L.; and Lerer, A. 2017.
\newblock Automatic differentiation in {PyTorch}.

\bibitem[{Qin et~al.(2020)Qin, Fang, Zhang, Liu, Wang, and
  Wang}]{qin_resizemix_2020}
Qin, J.; Fang, J.; Zhang, Q.; Liu, W.; Wang, X.; and Wang, X. 2020.
\newblock {ResizeMix}: {Mixing} {Data} with {Preserved} {Object} {Information}
  and {True} {Labels}.
\newblock \emph{arXiv:2012.11101 [cs]}.
\newblock ArXiv: 2012.11101.

\bibitem[{Shahroudy et~al.(2016)Shahroudy, Liu, Ng, and
  Wang}]{shahroudy_ntu_2016}
Shahroudy, A.; Liu, J.; Ng, T.-T.; and Wang, G. 2016.
\newblock {NTU} {RGB}+{D}: {A} {Large} {Scale} {Dataset} for {3D} {Human}
  {Activity} {Analysis}.
\newblock In \emph{2016 {IEEE} {Conference} on {Computer} {Vision} and
  {Pattern} {Recognition} ({CVPR})}, 1010--1019. Las Vegas, NV, USA: IEEE.
\newblock ISBN 978-1-4673-8851-1.

\bibitem[{Shi et~al.(2019)Shi, Zhang, Cheng, and Lu}]{shi_two-stream_2019}
Shi, L.; Zhang, Y.; Cheng, J.; and Lu, H. 2019.
\newblock Two-{Stream} {Adaptive} {Graph} {Convolutional} {Networks} for
  {Skeleton}-{Based} {Action} {Recognition}.
\newblock In \emph{2019 {IEEE}/{CVF} {Conference} on {Computer} {Vision} and
  {Pattern} {Recognition} ({CVPR})}, 12018--12027. Long Beach, CA, USA: IEEE.
\newblock ISBN 978-1-72813-293-8.

\bibitem[{Shi et~al.(2021)Shi, Zhang, Cheng, and Lu}]{shi_adasgn_2021}
Shi, L.; Zhang, Y.; Cheng, J.; and Lu, H. 2021.
\newblock {AdaSGN}: {Adapting} {Joint} {Number} and {Model} {Size} for
  {Efficient} {Skeleton}-{Based} {Action} {Recognition}.
\newblock \emph{arXiv:2103.11770 [cs]}.
\newblock ArXiv: 2103.11770.

\bibitem[{Si et~al.(2019)Si, Chen, Wang, Wang, and Tan}]{si_attention_2019}
Si, C.; Chen, W.; Wang, W.; Wang, L.; and Tan, T. 2019.
\newblock An {Attention} {Enhanced} {Graph} {Convolutional} {LSTM} {Network}
  for {Skeleton}-{Based} {Action} {Recognition}.
\newblock In \emph{2019 {IEEE}/{CVF} {Conference} on {Computer} {Vision} and
  {Pattern} {Recognition} ({CVPR})}, 1227--1236. Long Beach, CA, USA: IEEE.
\newblock ISBN 978-1-72813-293-8.

\bibitem[{Song et~al.(2017)Song, Lan, Xing, Zeng, and Liu}]{song_end--end_2017}
Song, S.; Lan, C.; Xing, J.; Zeng, W.; and Liu, J. 2017.
\newblock An {End}-to-{End} {Spatio}-{Temporal} {Attention} {Model} for {Human}
  {Action} {Recognition} from {Skeleton} {Data}.
\newblock \emph{Proceedings of the AAAI Conference on Artificial Intelligence},
  31(1).
\newblock Number: 1.

\bibitem[{Summers and Dinneen(2019)}]{summers_improved_2019}
Summers, C.; and Dinneen, M.~J. 2019.
\newblock Improved {Mixed}-{Example} {Data} {Augmentation}.
\newblock \emph{arXiv:1805.11272 [cs]}.
\newblock ArXiv: 1805.11272.

\bibitem[{Vemulapalli, Arrate, and Chellappa(2014)}]{vemulapalli_human_2014}
Vemulapalli, R.; Arrate, F.; and Chellappa, R. 2014.
\newblock Human {Action} {Recognition} by {Representing} {3D} {Skeletons} as
  {Points} in a {Lie} {Group}.
\newblock In \emph{2014 {IEEE} {Conference} on {Computer} {Vision} and
  {Pattern} {Recognition}}, 588--595. Columbus, OH, USA: IEEE.
\newblock ISBN 978-1-4799-5118-5.

\bibitem[{Verma et~al.(2019)Verma, Lamb, Beckham, Najafi, Mitliagkas,
  Courville, Lopez-Paz, and Bengio}]{verma_manifold_2019}
Verma, V.; Lamb, A.; Beckham, C.; Najafi, A.; Mitliagkas, I.; Courville, A.;
  Lopez-Paz, D.; and Bengio, Y. 2019.
\newblock Manifold {Mixup}: {Better} {Representations} by {Interpolating}
  {Hidden} {States}.
\newblock \emph{arXiv:1806.05236 [cs, stat]}.
\newblock ArXiv: 1806.05236.

\bibitem[{Wang et~al.(2014{\natexlab{a}})Wang, Liu, Wu, and
  Yuan}]{wang_learning_2014}
Wang, J.; Liu, Z.; Wu, Y.; and Yuan, J. 2014{\natexlab{a}}.
\newblock Learning {Actionlet} {Ensemble} for {3D} {Human} {Action}
  {Recognition}.
\newblock \emph{IEEE Trans. Pattern Anal. Mach. Intell.}, 36(5): 914--927.

\bibitem[{Wang et~al.(2014{\natexlab{b}})Wang, Nie, Xia, Wu, and
  Zhu}]{wang_cross-view_2014}
Wang, J.; Nie, X.; Xia, Y.; Wu, Y.; and Zhu, S.-C. 2014{\natexlab{b}}.
\newblock Cross-view {Action} {Modeling}, {Learning} and {Recognition}.
\newblock \emph{arXiv:1405.2941 [cs]}.
\newblock ArXiv: 1405.2941.

\bibitem[{Wojciech~Zaremba and Vinyals(2014)}]{Zaremba2014RecurrentNN}
Wojciech~Zaremba, I.~S.; and Vinyals, O. 2014.
\newblock Recurrent Neural Network Regularization.
\newblock \emph{ArXiv}, abs/1409.2329.

\bibitem[{Yan, Xiong, and Lin(2018)}]{yan_spatial_2018}
Yan, S.; Xiong, Y.; and Lin, D. 2018.
\newblock Spatial {Temporal} {Graph} {Convolutional} {Networks} for
  {Skeleton}-{Based} {Action} {Recognition}.
\newblock \emph{AAAI}, 32(1).
\newblock Number: 1.

\bibitem[{Ye et~al.(2020)Ye, Pu, Zhong, Li, Xie, and Tang}]{ye_dynamic_2020}
Ye, F.; Pu, S.; Zhong, Q.; Li, C.; Xie, D.; and Tang, H. 2020.
\newblock Dynamic {GCN}: {Context}-enriched {Topology} {Learning} for
  {Skeleton}-based {Action} {Recognition}.
\newblock In \emph{Proceedings of the 28th {ACM} {International} {Conference}
  on {Multimedia}}, 55--63. Seattle WA USA: ACM.
\newblock ISBN 978-1-4503-7988-5.

\bibitem[{Ye and Tang(2019)}]{ye2019skeleton}
Ye, F.; and Tang, H. 2019.
\newblock Skeleton-based action recognition with JRR-GCN.
\newblock \emph{Electronics Letters}, 55(17): 933--935.

\bibitem[{Ye et~al.(2019)Ye, Tang, Wang, and Liang}]{ye2019joints}
Ye, F.; Tang, H.; Wang, X.; and Liang, X. 2019.
\newblock Joints relation inference network for skeleton-based action
  recognition.
\newblock In \emph{2019 IEEE International Conference on Image Processing
  (ICIP)}, 16--20. IEEE.

\bibitem[{{Yong Du}, Wang, and Wang(2015)}]{yong_du_hierarchical_2015}
{Yong Du}; Wang, W.; and Wang, L. 2015.
\newblock Hierarchical recurrent neural network for skeleton based action
  recognition.
\newblock In \emph{2015 {IEEE} {Conference} on {Computer} {Vision} and
  {Pattern} {Recognition} ({CVPR})}, 1110--1118. Boston, MA, USA: IEEE.
\newblock ISBN 978-1-4673-6964-0.

\bibitem[{Yun et~al.(2012)Yun, Honorio, Chattopadhyay, Berg, and
  Samaras}]{yun_two-person_2012}
Yun, K.; Honorio, J.; Chattopadhyay, D.; Berg, T.~L.; and Samaras, D. 2012.
\newblock Two-person interaction detection using body-pose features and
  multiple instance learning.
\newblock In \emph{2012 {IEEE} {Computer} {Society} {Conference} on {Computer}
  {Vision} and {Pattern} {Recognition} {Workshops}}, 28--35. Providence, RI,
  USA: IEEE.
\newblock ISBN 978-1-4673-1612-5 978-1-4673-1611-8 978-1-4673-1610-1.

\bibitem[{Yun et~al.(2019)Yun, Han, Oh, Chun, Choe, and Yoo}]{yun_cutmix_2019}
Yun, S.; Han, D.; Oh, S.~J.; Chun, S.; Choe, J.; and Yoo, Y. 2019.
\newblock {CutMix}: {Regularization} {Strategy} to {Train} {Strong}
  {Classifiers} with {Localizable} {Features}.
\newblock \emph{arXiv:1905.04899 [cs]}.
\newblock ArXiv: 1905.04899.

\bibitem[{Zhang et~al.(2018)Zhang, Cisse, Dauphin, and
  Lopez-Paz}]{zhang_mixup_2018}
Zhang, H.; Cisse, M.; Dauphin, Y.~N.; and Lopez-Paz, D. 2018.
\newblock mixup: {Beyond} {Empirical} {Risk} {Minimization}.
\newblock \emph{arXiv:1710.09412 [cs, stat]}.
\newblock ArXiv: 1710.09412.

\bibitem[{Zhang et~al.(2019)Zhang, Lan, Xing, Zeng, Xue, and
  Zheng}]{zhang_view_2019}
Zhang, P.; Lan, C.; Xing, J.; Zeng, W.; Xue, J.; and Zheng, N. 2019.
\newblock View {Adaptive} {Neural} {Networks} for {High} {Performance}
  {Skeleton}-based {Human} {Action} {Recognition}.
\newblock \emph{arXiv:1804.07453 [cs]}.
\newblock ArXiv: 1804.07453.

\bibitem[{Zhang et~al.(2020)Zhang, Lan, Zeng, Xing, Xue, and
  Zheng}]{zhang_semantics-guided_2020}
Zhang, P.; Lan, C.; Zeng, W.; Xing, J.; Xue, J.; and Zheng, N. 2020.
\newblock Semantics-{Guided} {Neural} {Networks} for {Efficient}
  {Skeleton}-{Based} {Human} {Action} {Recognition}.
\newblock 1112--1121.

\bibitem[{Zhu et~al.(2016)Zhu, Lan, Xing, Zeng, Li, Shen, and
  Xie}]{zhu_co-occurrence_2016}
Zhu, W.; Lan, C.; Xing, J.; Zeng, W.; Li, Y.; Shen, L.; and Xie, X. 2016.
\newblock Co-{Occurrence} {Feature} {Learning} for {Skeleton} {Based} {Action}
  {Recognition} {Using} {Regularized} {Deep} {LSTM} {Networks}.
\newblock \emph{Proceedings of the AAAI Conference on Artificial Intelligence},
  30(1).
\newblock Number: 1.

\end{thebibliography}
}
	
\clearpage
\maketitle
\appendix
\section{Supplementary Material}
\subsection{Notes on the Data Modalities We Used}
In our experiments, we use single data modality (\ie the joint modality) for the ablation studies (Table~\ref{CAGVAG}, \ref{cm}, \ref{PM2}, \ref{tab:ablation-sm}, \ref{hard_gain}, \ref{GN} and \ref{smix}), evaluation of model complexity (Table~\ref{eff} and Figure~\ref{compare}) and visualizations (Figure \ref{se} and \ref{gain}). Four modalities, namely joint, bone, joint motion and bone motion are used only to report the final results of Ta-CNN and Ta-CNN+ in Table~\ref{N-UCLA}, \ref{tab:ntu60} and \ref{tab:ntu120}, which is the same as pervious works~\cite{cheng_skeleton-based_2020,chen_channel-wise_2021,ye_dynamic_2020,chen_multi-scale_2021}.

\subsection{Per-class Performance Improvement}
\begin{figure*}[t]
	\centering
	\includegraphics[width=0.98\textwidth]{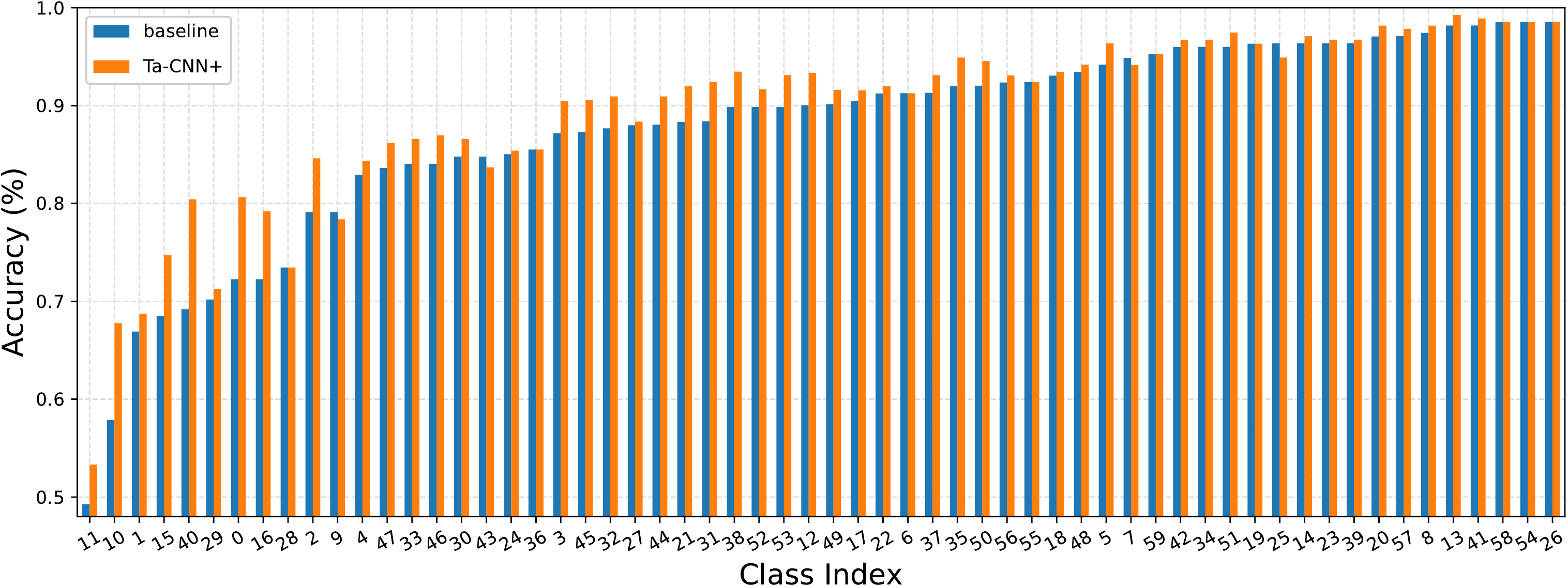} 
	\caption{Per-class accuracies of the strong baseline and Ta-CNN+ on the cross-subject benchmark of NTU RGB+D. The class indices are sorted by the accuracy of the baseline in ascending order.}
	\label{gain}
\end{figure*}

Figure~\ref{gain} shows the per-class accuracies of the strong baseline and Ta-CNN+. It is clear that Ta-CNN+ improves the baseline for most classes, especially the six classes, namely 0 (drink water), 2 (brushing teeth), 10 (reading), 15 (wear a shoe), 16 (take off a shoe) and 40 (sneeze). From Table~\ref{hard_gain}, we can find the classes improved incredibly are also the most difficult classes for the baseline to recognize. These classes share two common characteristics. One is the high similarity between the actions, \eg wear a shoe and take off a shoe. The other is the subtle movement involved such as reading. Equipped with CAG and VAG, Ta-CNN+ enhances the cross-channel features in multiple low-dimensional subspaces and virtual parts. As a result, Ta-CNN+ is capable of identifying the nuanced actions, away from their similar classes.
\begin{table}[ht]
	\small
	\centering
	
	\begin{tabular}{c|c|c}
		\toprule
		\multicolumn{1}{c|}{Class} & Baseline & Ta-CNN+ \\
		\midrule
		drink water & 72.3  & 80.7 $(\uparrow 8.4)$  \\
		brushing teeth & 79.1  & 84.6 $(\uparrow 5.5)$  \\
		reading & 57.9  & 67.8 $(\uparrow 9.9)$  \\
		wear a shoe & 68.5  & 74.7 $(\uparrow 6.2)$  \\
		take off a shoe & 72.3  & 79.2 $(\uparrow 6.9)$  \\
		sneeze/cough & 69.2  & 80.4 $(\uparrow 11.2)$  \\
		\bottomrule
	\end{tabular}%
	\caption{Accuracy gains by Ta-CNN+ for 6 classes on the cross-subject benchmark of NTU RGB+D.}
	\label{hard_gain}%
\end{table}%

\subsection{Impact of the Groups in CAG and VAG}
To show the impact of the hyper-parameter $n$ which regulates the number of groups in CAG and VAG, we report the performance of various combinations of CAG and VAG in Table~\ref{GN}. The top-2 combinations are (10,6) and (10,10). This result is consistent with the reality. For the CAG module, $n=10$ means the feature space is divided into 10 3-dimensional subspaces, which is analogous to the $x,y,z$ input coordinates. For the VAG module, $n=10$ or $n=6$ means three or five joints are chosen as a part, which is comparable to key parts such as hand (3 joints) and arm (5 joints) in NTU RGB+D~\cite{shahroudy_ntu_2016}.

\begin{table}[ht]
	\centering
	
	\begin{tabular}{c|c|c|c}
		\toprule
		&  VAG (6)     &  VAG (8)    & VAG (10) \\
		\midrule
CAG (6)		&  88.5  & 88.5   &  88.6\\
CAG (8)		&  88.4  & 88.5 & 88.5 \\
CAG (10)		&  \textbf{88.8}  & 88.4   & \textbf{88.7} \\
		\bottomrule
	\end{tabular}%
	\caption{Performance of various combinations of CAG and VAG. The numbers, namely 6, 8 and 10, mean the hyper-parameter $n$ regulating the number of groups (performed on X-sub benchmark of NTU RGB+D).}
	\label{GN}%
\end{table}%
\begin{table}[ht]
	\small
	\centering
	\begin{tabular}{l|c|c|c|c}
		\toprule
		\multicolumn{1}{l|}{\multirow{2}[4]{*}{Modality}} & \multicolumn{2}{c|}{w/o SkeletonMix} & \multicolumn{2}{c}{w/ SkeletonMix} \\
		\cmidrule{2-5}          & \multicolumn{1}{l|}{X-sub} & \multicolumn{1}{l|}{X-view} & \multicolumn{1}{l|}{X-sub} & \multicolumn{1}{l}{X-view} \\
		\midrule
		Joint & 88.7  & 93.3  & 88.8  & 93.6  \\
		Bone  & 87.1  & 93.0  & 88.0  & 93.2  \\
		Joint Motion & 84.9  & 90.1  & 85.8  &  90.5 \\
		Bone Motion  & 84.2  & 89.2  & 84.8  & 89.8  \\
		\bottomrule
	\end{tabular}%
	\caption{Performance of SkeletonMix when trained using various data modalities on NTU RGB+D.}
	\label{PM3}%
\end{table}%
\begin{table}[!ht]
	\centering
	\begin{tabular}{c|c|c|c|c|c}
		\toprule
		$\alpha$     & \multicolumn{1}{l|}{$1/2$} & \multicolumn{1}{l|}{$1/4$} & \multicolumn{1}{l|}{$1/8$} & \multicolumn{1}{l|}{$1/16$} & \multicolumn{1}{l}{$1/32$} \\
		\midrule
		Acc (\%)  &  82.0  &  82.0   &  82.1   &  \textbf{82.4} &  82.3\\
		\bottomrule
	\end{tabular}%
	\caption{Impacts of different $\alpha$ in SkeletonMix on X-sub benchmark of NTU RGB+D 120.}
	\label{smix}%
\end{table}%

\subsection{SkeletonMix}

We conduct supplementary experiments to validate the SkeletonMix strategy on the NTU RGB+D dataset. As shown in Table~\ref{PM3}, SkeletonMix improves the recognition accuracy, no matter what data modality is used as input, which is consistent with the experiments on NTU RGB+D 120 (Table~\ref{PM2}). In addition, we explore the impact of different $\alpha$. As summarized in Table~\ref{smix}, the accuracy of the model first increases and then decreases as $\alpha$ decreases, and the best result is achieved with $\alpha=1 / 16$.

\subsection{Architecture of Ta-CNN}
\begin{table*}[!ht]
	\centering
	\begin{tabular}{c|l|l|l|c|c}
		\toprule
		\multicolumn{1}{c|}{Module} & \multicolumn{1}{l|}{Input} & \multicolumn{1}{l|}{Output} & \multicolumn{1}{l|}{Ta-CNN}& \multicolumn{1}{c|}{Param. (M)}& \multicolumn{1}{c}{GFLOPs} \\
		\midrule
		\multirow{6}[12]{*}{CAG*2} & $3\times64\times25$  & $64\times64\times25$   & $conv 1\times1,64,BN, ReLU$& $0.000$&$0.001$ \\
		\cmidrule{2-3}          &  $64\times64\times25$      &   $30\times64\times25$     &  $ conv 1\times1,30$& $0.004$& $0.006$\\
		\cmidrule{2-6}          &   $30\times64\times25$    &   $30\times64\times25$    &  $SE, r=1$&$0.004$& $0.000$\\
		\cmidrule{2-6}          &   $30\times64\times25$    &   $30\times64\times25$    &  $ conv3\times1,30,group 10$& $0.001$& $0.001$\\
		\cmidrule{2-3}          &   $30\times64\times25$    &   $30\times64\times25$    &  $conv1\times1 ,30,group 5$& $0.000$& $0.001$\\
		\cmidrule{2-6}          &    $30\times64\times25$   &   $32\times64\times25$    &  $conv1\times1 ,32$& $0.002$& $0.003$\\
		\midrule
		Transpose &    $32\times64\times25$   &   $25\times64\times32$    &  $transpose(3,2,1)$& $0.000$& $0.000$\\
		\midrule
		\multirow{6}[12]{*}{VAG*2} &   $25\times64\times32$    &   $30\times64\times32$    &  $ conv1\times1,30$& $0.002$& $0.003$\\
		\cmidrule{2-6}          &   $30\times64\times32$    &  $30\times64\times32$     & $SE, r=1$& $0.004$& $0.000$\\
		\cmidrule{2-6}          &   $30\times64\times32$    &  $30\times64\times32$     & 	$conv3\times3 ,30,group 10$ &$0.002$& $0.003$ \\
		\cmidrule{2-3}          &    $30\times64\times32$   & $30\times64\times32$      &  $conv1\times1 ,30,group 5$& $0.000$& $0.001$\\
		\cmidrule{2-6}          & $30\times64\times32$      &   $32\times64\times32$      & $conv1\times1 ,32$ & $0.002$& $0.004$\\
		\cmidrule{2-3}          &    $32\times64\times32$    &  $64\times16\times8$      & $Maxpool,conv3\times3 ,64,Maxpool$& $0.037$& $0.019$\\
		\midrule
		Concat &  $64\times16\times8$       &   $128\times16\times8$      &  $concat(1),Dropout(0.5)$ & $0.000$& $0.000$\\
		\midrule
		\multirow{2}[6]{*}{Convs} &   $128\times16\times8$     &   $128\times8\times4$     &  $conv3\times3 ,128,Maxpool,ReLU,Dropout(0.5)$& $0.147$& $0.019$\\
		\cmidrule{2-6}          &   $128\times8\times4$      &     $256\times4\times2$    &  $conv3\times3 ,256,Maxpool,ReLU$& $0.295$& $0.009$\\
		\midrule
		Mean          &    $256\times4\times2$     &    $256\times1\times2$     &  $mean(2)$& $0.000$& $0.000$\\
		\midrule
		Maxout &   $256\times1\times2\times M$       &  $256\times1\times2$        & $maxout$ & $0.000$& $$0.000$$\\
		\midrule
		FC    &  $256\times1\times2$     &   $classes$      & $Flatten,Dropout(0.5),512\times classes fc$& $0.031$& $0.000$ \\
		\bottomrule
	\end{tabular}%
	\caption{The detailed architecture of Ta-CNN. Maxout is applied dynamically to $M$ varying number of persons.}
	\label{model}%
\end{table*}%
Table~\ref{model} lists the detailed configuration of each module in Ta-CNN, including the data shape, parameters and GFLOPs. CAG and VAG are repeated twice for the two input streams, \ie the skeleton sequence and the skeleton motion respectively. For samples involving multiple persons, we adopt a dynamic fusion strategy. Thus the real GFLOPs is approximately proportional to the number of input persons and may vary for different samples. The GFLOPs reported in Table~\ref{eff} is the average value of the samples in NTU RGB+D and NTU RGB+D 120.

\end{document}